\documentclass[10pt,twocolumn]{article}

\usepackage[margin=0.72in]{geometry}
\usepackage[T1]{fontenc}
\usepackage[utf8]{inputenc}
\usepackage{amsmath,amssymb,amsfonts}
\usepackage{graphicx}
\usepackage{booktabs}
\usepackage{array}
\usepackage{float}
\usepackage{placeins}
\usepackage{url}
\usepackage{microtype}
\usepackage{authblk}
\usepackage[hidelinks]{hyperref}
\graphicspath{{figures/}}

\title{Execution Realism and Reproducibility in LLM-Based Trading Systems:\\
A Systematic Scoping Review and Evidence Audit}

\author[1]{Junyi Yao\thanks{ORCID: 0009-0005-6375-2503; e-mail: j.yao@wustl.edu}}
\author[1]{Zihao Zheng\thanks{E-mail: z.zihaogary@wustl.edu}}
\author[2]{Baichuan Li\thanks{Corresponding author; e-mail: baichuanl@smu.edu}}
\affil[1]{Washington University in St. Louis, St. Louis, MO 63130, USA}
\affil[2]{Department of Operations Research and Engineering Management, Southern Methodist University, Dallas, TX 75275, USA}

\date{}

\begin{document}
\raggedbottom
\maketitle

\begin{abstract}
Execution realism remains weakly standardized in research on large-language-model-based trading systems, limiting comparison and reproduction across backtests, simulations, and portfolio benchmarks. This article presents a PRISMA-ScR systematic scoping review with a nested execution-reproducibility evidence audit. Eight reproducible query families returned 796 query-level records; DOI/title deduplication left 687 unique records, 101 reports were sought for retrieval, and 59 full texts were recovered through direct and fallback open-access routes. Fifty-three reports are provisionally eligible for expanded evidence charting, subject to reconciliation of two blind author-validation packets. The charting framework separately evaluates point-in-time controls, temporal splits, held-out evaluation, trading costs and turnover, execution semantics, universe construction, architecture reporting, and artifact availability. Direct-trading, portfolio or alpha-construction, and benchmark studies are retained as distinct descriptive subgroups rather than pooled as a common performance estimand. Search responses, retrieval attempts, file hashes, evidence excerpts, screening decisions, and coding worksheets are archived with the review materials. Final field-level counts are intentionally withheld until author validation is complete.
\end{abstract}


\section{Introduction}

Execution realism has become a central comparability problem in research on large-language-model-based trading. Large language models (LLMs) now support finance-oriented retrieval, reasoning, decision support, and autonomous workflows. Recent systems gather heterogeneous information, maintain memory, deliberate over market scenarios, and coordinate analyst, trader, and risk-management roles \cite{survey2024,tradinggpt2023,stockagent2024,tradingagents2024}. Their architectural appeal is clear: financial decisions combine prices, filings, news, macroeconomic signals, and portfolio constraints under changing regimes \cite{finmem2023,finagent2024,cryptotrade2024}.

Evaluation practice has not matured at the same pace. A reported return depends not only on the decision model but also on when information became available, when orders could execute, whether costs and slippage were charged, how the asset universe was formed, how out-of-sample periods were selected, and whether code, prompts, model versions, and data snapshots were documented. Empirical finance has long shown that data snooping, backtest overfitting, and omitted trading costs can inflate apparent performance \cite{white2000reality,sullivan1999datasnooping,bailey2014deflated,korajczyk2004momentum}. LLM workflows add further provenance and version-control risks because language, retrieval, memory, and tool use may each alter the effective information set.

Existing surveys provide valuable architecture-centered maps of financial LLMs and trading agents \cite{survey2024,agentictrading2026}. The present study addresses a different question: among LLM-based systems connected to trading, portfolio construction, alpha research, market simulation, or trade-relevant decisions, which execution and reproducibility assumptions are recoverable from the public record, and where do reporting gaps prevent economic interpretation or independent reproduction? We answer this question through a bounded systematic scoping review reported with PRISMA-ScR and PRISMA-S guidance, together with a nested study-level evidence audit. Unlike architecture-centered evidence maps, the unit of analysis here is the public evidence supporting each execution assumption: every code is tied to a study-level record and recoverable passage, and heterogeneous system types are synthesized only within descriptive subgroups.

The contributions are as follows:
\begin{enumerate}
\item We define and archive a reproducible, date-bounded search, deduplication, screening, retrieval, and version-reconciliation workflow for trade-relevant LLM studies.
\item We chart point-in-time discipline, temporal splits, held-out evaluation, transaction costs and turnover, execution semantics, universe construction, and artifact availability using explicit Y/P/NR/N/A rules and source anchors.
\item We separate machine-assisted retrieval and passage detection from final human eligibility and coding decisions, and provide independent blind validation worksheets for two authors.
\item We organize the evidence into direct-trading, portfolio or alpha-construction, and benchmark or decision-evaluation subgroups without treating heterogeneous outcomes as a common meta-analytic estimand.
\item We distinguish recoverability from study quality and predictive validity, avoiding a composite score that would pool inapplicable criteria across heterogeneous study types.
\item We derive a minimum reporting checklist for future authors, reviewers, and benchmark designers.
\end{enumerate}

\section{Related Work: From Agent Designs to Evaluation Assumptions}

\subsection{Financial LLM Infrastructure and Review Scope}

Financial language models and agent platforms provide the infrastructure on which trading systems are built. BloombergGPT and FinGPT emphasize finance-specific corpora and tasks \cite{bloomberggpt2023,fingpt2023}; FinanceBench, FinBen, and related evaluations span question answering, extraction, forecasting, risk, and decision support \cite{llmfinance2023,finllms2024,financebench2023,finben2024}. FinRobot connects models to retrieval, analytical tools, and workflow orchestration \cite{finrobot2024}. Adjacent work on factor decomposition and bankruptcy prediction from filings shows how structured market variables and textual evidence are increasingly combined \cite{han2026interpretable,zhang2026bankruptcy}.

These studies motivate, but do not automatically enter, the primary evidence set. Financial question answering, sentiment analysis, retrieval, or forecasting is included only when the study establishes an explicit path from model output to a tradable decision, portfolio, market simulation, or evaluation benchmark. That boundary distinguishes general financial NLP from claims whose interpretation depends on execution assumptions \cite{chatgptforecast2023,sentimenttrading2024}.

\subsection{Architecture-Centered Surveys and Systems}

Prior surveys map trading-agent architectures, inputs, and reported performance \cite{survey2024,agentictrading2026}. Memory and reflection systems include FinMem and reflective cryptocurrency agents \cite{finmem2023,cryptotrade2024}. Multi-agent systems decompose work among specialized roles or simulated participants \cite{tradinggpt2023,stockagent2024,tradingagents2024,alphaagents2025,contesttrade2025,expertteams2026,alphacrafter2026}. Multimodal and tool-augmented systems fuse text, market data, visual signals, routing, and computational traces \cite{finagent2024,mmdrex2025,qrafti2026}. Alpha-mining and hybrid policy systems place LLMs closer to factor discovery, signal generation, and quantitative research automation \cite{alphagpt2023,quantagent2024,stockgpt2024,alphaagent2025,finrldeepseek2025,hubble2026}.

Architecture families explain design intent but do not make reported returns comparable. Memory, debate, and multimodal systems can all report performance under materially different information sets, fill rules, costs, universes, and validation windows. The present review therefore treats architecture as a study characteristic and execution-reproducibility reporting as the primary charting target.

More generally, persistent agent memory creates an audit problem because stale or faulty records can propagate into later reasoning and actions. Execution-provenance research organizes traces and evidence dependencies for process-level accountability \cite{wang2026agenttracestotrust}; provenance-aware shared memory additionally uses lineage, access constraints, and action-risk gates \cite{wang2026mapgraph}. Transactional belief commits and dependency-guided rollback illustrate complementary controls for preventing unsafe state propagation and selectively repairing affected actions \cite{li2026memtx,yu2026rollbackrepair}. These studies provide background for the review's emphasis on recoverable execution records, but they are not treated as primary evidence about trading performance.

Outside finance, agentic reasoning, program-guided execution, and reusable policy routines illustrate broader ways to decompose long-horizon decisions into explicit components \cite{rma2026,proto2021,zhao2021augmenting}. These works motivate architecture descriptors but do not provide evidence about financial execution realism; accordingly, they are treated only as background and are excluded from the primary trading-study audit.

Likewise, work on grounded neural representations and probabilistic object tracking illustrates how upstream perception modules can expose structured intermediate representations \cite{zhao2024grounding,zhao2022tracking}. Such perception research is relevant only to the modular-systems context; it is not used to support claims about trading performance, costs, or reproducibility.

\subsection{Benchmarks, Portfolios, and Execution Realism}

InvestorBench, AI-Trader, PortBench, Look-Ahead-Bench, and LLM market simulations move the literature toward structured financial decision evaluation \cite{investorbench2024,aitrader2025,portbench2026,lookaheadbench2026,llmtrade2025}. Point-in-time and event-driven designs are particularly important because they reduce retrospective leakage. Portfolio settings add allocation, correlation, turnover, and regime sensitivity, while calibrated technical strategies show that parameter and friction sensitivity matter even without an LLM \cite{lin2026volume}. Recent audits of foundation-model evaluation likewise show that aggregate benchmark results can conceal regime-dependent failures \cite{wang2026timeseriesfoundationmodel}. Related work embeds foundation-model predictions in a structurally constrained discrete-choice model, illustrating how predictive components can be coupled to explicit behavioral and economic guarantees \cite{wang2026embeddingfoundationmodelpredictions}. Adversarial news and live or simulated evaluations further test whether apparent skill survives temporal separation and manipulation \cite{adversarialnews2026,aitrader2025,llmtrade2025}.

Execution assumptions are often mundane but decisive. Same-close fills can overstate performance when a signal uses closing information; high turnover can erase gross returns after commissions, spread, and slippage; future constituents introduce survivorship bias; and closed model endpoints without version records impede reproduction. These observations motivate the evidence fields and reporting checklist used in this review.

\section{Systematic Scoping-Review Protocol}

\subsection{Review Design and Question}

This study is designed as a systematic scoping review with a nested execution-reproducibility evidence audit. Reporting follows PRISMA-ScR \cite{tricco2018prismascr}, the PRISMA 2020 flow logic \cite{prisma2020}, and PRISMA-S search-reporting principles \cite{rethlefsen2021prismas}. A scoping design is appropriate because the literature spans direct trading agents, portfolio and alpha-research systems, market simulations, and decision benchmarks with incompatible outcomes, markets, horizons, and evaluation protocols. A pooled meta-analysis would therefore imply a common estimand that the source literature does not provide.

The review asks: among LLM-based or LLM-agent systems connected to trading, portfolio construction, alpha research, market simulation, or trade-relevant financial decisions, which execution and reproducibility assumptions are publicly recoverable, and where do reporting gaps prevent economic interpretation or independent reproduction?

The publication window is January~1,~2023 through May~30,~2026. English-language journal articles, conference papers, accessible preprints, and technical reports are eligible. Secondary reviews are retained for background and citation searching but are not coded as primary evidence. The review was not prospectively registered. A dated protocol reconstructed and frozen during the post-rejection redesign is supplied with the review materials; this absence of prospective registration is treated as a limitation rather than obscured.

\subsection{Information Sources and Search}

Eight query families cover large language models, LLM agents, multi-agent trading, portfolio management, alpha mining, financial backtesting, and quantitative trading. The machine-readable search script records the exact strings, date bounds, returned metadata, query-level counts, and OpenAlex identifiers. The search was rerun on October~4,~2026 while preserving the prespecified publication cutoff of May~30,~2026. Because several broad query families produced very large result sets, the bounded search retained the first 100 relevance-ranked OpenAlex records per query family. This cap makes the review reproducible but introduces ranking-based selection risk; the review is systematic within the declared bounds rather than an exhaustive census. Citation searching from retained studies and prior financial-LLM reviews supplements the database search. Source versions are reconciled by DOI and normalized title, with the most complete accessible report retained.

The bounded electronic search archived 796 query-level records. DOI/title deduplication removed 109 records, leaving 687 unique records for title and abstract screening. The raw JSON response, deduplicated CSV, search summary, scripts, and retrieval logs are included in the review materials so the search can be rerun or audited.

\subsection{Eligibility and Screening}

Eligibility is based on population, concept, and context. The population is an LLM, foundation model, or LLM-agent system. The concept is an empirical trading, portfolio, alpha-factor, market-simulation, or trade-relevant benchmark evaluation. The context is securities, cryptocurrencies, investment portfolios, or financial decisions with an explicit path to tradable action. Pure financial NLP, generic forecasting without a decision path, non-investment trading contexts, tutorials, editorials, and secondary reviews are excluded from the primary evidence set.

Records are screened in two stages. Title/abstract screening applies cascading, fixed rules: concept screening first requires an LLM/agent term and a finance/trading term in the title or indexed abstract; title-level scope refinement then requires an LLM/agent and trade-relevant term in the title; and metadata/abstract triage removes secondary reviews, non-investment trading contexts, and conceptual or advisory studies without an empirical trade-evaluation claim. Together these rules provisionally exclude 586 records: 280 without a finance/trading concept, 28 without an LLM/agent concept, 260 at title-level scope refinement, eight secondary reviews or perspectives, seven non-investment trading records, and three conceptual or advisory records. The remaining 101 reports proceed to retrieval. Full-text screening uses E5--E10 and inclusion codes I1--I3. Machine-assisted triage is used only to prioritize retrieval and expose potentially relevant passages; it does not constitute a final eligibility decision. Direct PDF links, arXiv DOI reconstruction, SSRN delivery links, DOI landing pages, and open-access page links are attempted in sequence. Every retrieval attempt and successful file hash is logged.

The audit measures reporting recoverability, not predictive validity, study quality, or risk of bias. A Y code means that an execution-relevant assumption can be reconstructed or stress-tested from public evidence; it does not imply that the assumption is economically realistic or that the reported strategy is profitable. Conversely, NR records absence of recovered evidence rather than proof that the underlying implementation omitted the feature. This distinction prevents the charting scheme from becoming an unsupported composite quality score.

\begin{table}[H]
\caption{Provisional PRISMA-ScR Screening Summary}
\label{tab:prisma_screening_summary}
\centering
\begin{tabular}{p{0.72\columnwidth}r}
\toprule
Stage & Count \\
\midrule
Records retrieved across eight query families & 796 \\
Duplicate DOI/title records removed & 109 \\
Deduplicated records screened & 687 \\
Excluded by cascading title/abstract rules & 586 \\
Reports sought for retrieval & 101 \\
Reports not retrieved after fallback attempts & 42 \\
Full-text reports assessed & 59 \\
Low empirical signal, pending exclusion confirmation & 6 \\
Provisionally eligible for evidence charting & 53 \\
\bottomrule
\end{tabular}
\par\vspace{2pt}
\begin{minipage}{0.94\columnwidth}\footnotesize Exclusions comprise 280 records without a finance/trading concept, 28 without an LLM/agent concept, 260 failing title-level scope refinement, eight secondary reviews or perspectives, seven non-investment trading records, and three conceptual/advisory records without an empirical trade-evaluation claim. Final inclusion remains subject to two-author reconciliation.\end{minipage}
\end{table}

\begin{figure*}[!t]
\centering
\includegraphics[width=0.58\textwidth]{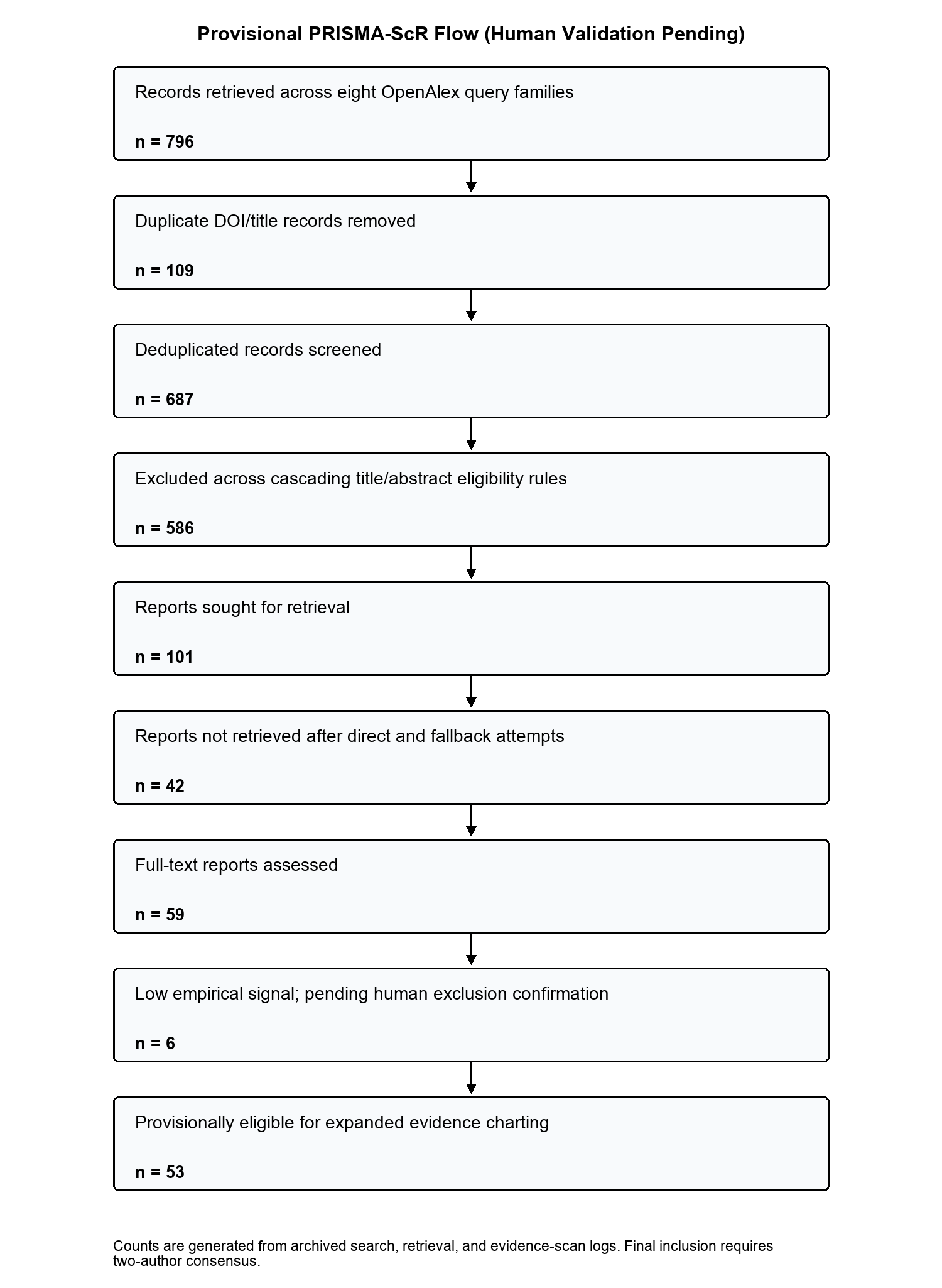}
\caption{Provisional PRISMA-ScR flow. Counts are generated from archived search, deduplication, retrieval, and evidence-scan logs. Final inclusion requires reconciliation of the two blind author-validation packets.}
\label{fig:prisma_flow}
\end{figure*}

\subsection{Data Charting and Audit Variables}

For each provisionally eligible report, the charting form records study role, market and task scope, temporal split or point-in-time controls, held-out evaluation, cost or turnover treatment, execution semantics, universe construction, system-architecture reporting, and artifact availability. ``Y'' denotes explicit evidence detailed enough to reconstruct or stress-test the field, ``P'' denotes partial or indirect recoverability, ``NR'' denotes that no supporting evidence was recovered, and ``N/A'' denotes a field genuinely outside the study's scope. Architecture is charted as a descriptive-reporting variable, not as evidence of trading validity; this prevents architecture clarity from being asserted without direct coding.

The extraction pipeline first identifies candidate passages using fixed regular expressions and stores the surrounding text. Two blind validation packets then present the study, field, and evidence excerpt without the machine-proposed code. Each author assigns Y, P, NR, or N/A and supplies a page or section anchor. Disagreements are reconciled by consensus. Until this reconciliation is complete, generated entries and aggregate counts remain provisional and are not used as final inferential claims.

\begin{table*}[!t]
\caption{Evaluation Assumptions That Determine Whether LLM Trading Results Are Economically Interpretable}
\label{tab:evaluation_risks}
\centering
\begin{tabular}{p{0.18\textwidth}p{0.34\textwidth}p{0.36\textwidth}}
\toprule
Dimension & Coding Question & Why It Matters \\
\midrule
Point-in-time data & Are all inputs available before the decision time? & Prevents future information from entering prompts, retrieval, features, or benchmark labels. \\
Temporal split & Are train, validation, tuning, and test periods explicitly dated? & Makes out-of-sample claims inspectable and reduces hidden tuning leakage. \\
Execution timing & When does a signal become an executable order? & Distinguishes same-close, next-open, next-close, and delayed execution assumptions. \\
Trading frictions & Are commissions, spreads, slippage, and turnover reported? & Determines whether active-strategy gains survive realistic implementation costs. \\
Universe construction & Is the asset universe fixed using information available at the time? & Reduces survivorship and hindsight bias in stock or portfolio selection. \\
Artifact release & Are code, prompts, data identifiers, and model versions documented? & Supports replication despite model drift, prompt sensitivity, and changing APIs. \\
\bottomrule
\end{tabular}
\end{table*}

\begin{table*}[!t]
\caption{Coding Rulebook for Execution-Reproducibility Fields}
\label{tab:coding_rulebook}
\centering
\scriptsize
\begin{tabular}{p{0.17\textwidth}p{0.24\textwidth}p{0.24\textwidth}p{0.24\textwidth}}
\toprule
Field & Y & P & NR \\
\midrule
Split/PiT & Explicit dated train, validation, and test periods or a clear point-in-time protocol. & Dates, periods, or temporal controls are partially recoverable, but not enough for full reconstruction. & No recoverable split dates or point-in-time controls in the public manuscript or linked artifacts. \\
Held-out & A temporally or otherwise independent test set is explicitly identified and separated from training, tuning, or prompt development. & Evaluation separation is suggested, but the independence of the reported test is not fully reconstructable. & No recoverable held-out evaluation protocol. \\
Costs & Transaction costs, spreads, slippage, or turnover are explicitly modeled or reported. & Costs or turnover are mentioned, or can be approximately inferred, but the implementation is incomplete. & No recoverable cost, slippage, or turnover treatment. \\
Execution & Signal formation and order timing are specified clearly enough to distinguish same-period and delayed execution. & Timing is implied by the environment or backtest description but not implementation-ready. & No recoverable execution-timing semantics. \\
Universe & Asset universe and inclusion rules are specified with enough detail to assess survivorship or selection bias. & Universe is named but constituent timing or selection rules remain partial. & No recoverable universe-construction rule. \\
Architecture & Agents, modules, roles, information flow, and decision interfaces are specified sufficiently to reconstruct the system design. & The architecture family or major components are named, but interfaces or workflow details remain incomplete. & No recoverable system-architecture description. \\
Artifacts & Code, prompts, data identifiers, model versions, or reproducibility package are released. & Some artifacts are released or described, but not enough for independent replication. & No recoverable reproducibility artifacts. \\
\bottomrule
\end{tabular}
\par\vspace{2pt}
\begin{minipage}{0.94\textwidth}\footnotesize Ambiguous cases are coded conservatively. A field is marked Y only when the manuscript gives enough information for an informed reader to reconstruct or stress-test the assumption. N/A is reserved for a field genuinely outside a study's scope and is never used merely because reporting is absent.\end{minipage}
\end{table*}

\subsection{Synthesis}

The synthesis is descriptive rather than meta-analytic. Final reporting will provide the PRISMA-ScR flow, study characteristics, field-level recoverability counts, direct-trading versus portfolio/alpha versus benchmark subgroups, and study-level caveats. No composite quality ranking is constructed because field applicability differs across study types and recoverability is not equivalent to empirical validity.

\begin{table*}[!t]
\caption{Expanded PRISMA-ScR Evidence Matrix, Part I of III (Provisional Pending Two-Author Validation)}
\label{tab:prisma_expanded_matrix}
\centering
\scriptsize
\begin{tabular}{p{0.30\textwidth}p{0.14\textwidth}ccccccc}
\toprule
Study & Subgroup & Split/PiT & Held-out & Costs & Exec. & Univ. & Arch. & Art. \\
\midrule
A Multi-agent System Based On LLM For Trading Financial Assets \cite{prismaft002} & Direct trading/simulation & NR & NR & P & P & NR & Y & NR \\
A Reflective LLM-based Agent to Guide Zero-shot Cryptocurrency Trading \cite{prismaft003} & Direct trading/simulation & P & Y & P & NR & NR & Y & P \\
Adversarial News and Lost Profits: Manipulating Headlines in LLM-Driven Algorithmic Trading \cite{adversarialnews2026} & Direct trading/simulation & Y & Y & P & Y & NR & Y & Y \\
Agentic AI for Clustering, Relationship Discovery, and Semantic Trading in Prediction Markets \cite{prismaft013} & Direct trading/simulation & NR & NR & P & Y & NR & Y & NR \\
Agentic AI in Commodity Trading: A Comparative Simulation Study \cite{prismaft015} & Direct trading/simulation & NR & NR & Y & NR & NR & Y & NR \\
AgenticAITA: A Proof-Of-Concept About Deliberative Multi-Agent Reasoning for Autonomous Trading Systems \cite{prismaft017} & Direct trading/simulation & Y & NR & P & Y & NR & Y & NR \\
Alpha-GPT: Human-AI Interactive Alpha Mining for Quantitative Investment \cite{alphagpt2023} & Portfolio/alpha & Y & Y & P & Y & Y & Y & Y \\
AlphaAgent: LLM-Driven Alpha Mining with Regularized Exploration to Counteract Alpha Decay \cite{alphaagent2025} & Portfolio/alpha & P & Y & P & Y & Y & Y & Y \\
An End-To-End LLM Enhanced Trading System \cite{prismaft026} & Direct trading/simulation & NR & NR & NR & P & NR & P & Y \\
Analysis of the Application of Large Language Models in Three Major Fields: Trading and Portfolio Management, Financial Risk Management, and Financial Text Mining \cite{prismaft028} & Portfolio/alpha & P & NR & NR & P & NR & P & NR \\
Can ChatGPT Forecast Stock Price Movements? Return Predictability and Large Language Models \cite{chatgptforecast2023} & Direct trading/simulation & Y & Y & Y & Y & NR & Y & P \\
CN-Buzz2Portfolio: A Chinese-Market Dataset and Benchmark for LLM-Based Macro and Sector Asset Allocation from Daily Trending Financial News \cite{prismaft041} & Benchmark/decision & P & P & Y & NR & Y & Y & P \\
Constructing a Portfolio Optimization Benchmark Framework for Evaluating Large Language Models \cite{prismaft042} & Benchmark/decision & NR & NR & NR & NR & NR & P & Y \\
CryptoTrade: A Reflective LLM-based Agent to Guide Zero-shot Cryptocurrency Trading \cite{cryptotrade2024} & Direct trading/simulation & P & Y & P & NR & NR & Y & Y \\
Decision-informed Neural Networks with Large Language Model Integration for Portfolio Optimization \cite{prismaft046} & Portfolio/alpha & Y & Y & NR & NR & P & Y & P \\
Evolutionary Factor Searching for Sparse Portfolio Optimization Using Large Language Models \cite{prismaft051} & Portfolio/alpha & Y & Y & Y & Y & P & Y & Y \\
Exploring LLM Cryptocurrency Trading Through Fact-Subjectivity Aware Reasoning \cite{prismaft053} & Direct trading/simulation & P & Y & NR & NR & NR & Y & Y \\
Exploring Sentiment Manipulation by LLM-Enabled Intelligent Trading Agents \cite{prismaft054} & Direct trading/simulation & Y & Y & P & Y & P & Y & NR \\
\bottomrule
\end{tabular}
\par\vspace{2pt}
\begin{minipage}{0.94\textwidth}\footnotesize Y = explicit evidence recovered; P = partial evidence; NR = not recovered. Entries remain provisional until the two blind validation packets are reconciled.\end{minipage}
\end{table*}
\begin{table*}[!t]
\caption{Expanded PRISMA-ScR Evidence Matrix, Part II of III (Provisional Pending Two-Author Validation)}
\centering
\scriptsize
\begin{tabular}{p{0.30\textwidth}p{0.14\textwidth}ccccccc}
\toprule
Study & Subgroup & Split/PiT & Held-out & Costs & Exec. & Univ. & Arch. & Art. \\
\midrule
Financial News-Driven LLM Reinforcement Learning for Portfolio Management \cite{prismaft056} & Portfolio/alpha & NR & NR & P & NR & NR & Y & NR \\
Fine-Tuning Large Language Models for Stock Return Prediction Using Newsflow \cite{prismaft057} & Direct trading/simulation & Y & Y & NR & P & P & Y & Y \\
FinLLM-B: When Large Language Models Meet Financial Breakout Trading \cite{prismaft058} & Direct trading/simulation & P & P & NR & NR & NR & P & NR \\
FinPos: A Position-Aware Trading Agent System for Real Financial Markets \cite{prismaft059} & Direct trading/simulation & Y & P & P & Y & Y & Y & P \\
FinRL-DeepSeek: LLM-Infused Risk-Sensitive Reinforcement Learning for Trading Agents \cite{finrldeepseek2025} & Direct trading/simulation & P & P & NR & NR & NR & Y & Y \\
FLAG-TRADER: Fusion LLM-Agent with Gradient-based Reinforcement Learning for Financial Trading \cite{prismaft062} & Direct trading/simulation & P & NR & P & P & NR & Y & P \\
From Knowing to Doing: A Memory-Controlled Benchmark for LLM Trading Agents on Stock Markets \cite{prismaft064} & Benchmark/decision & P & P & P & Y & NR & Y & P \\
From Text to Returns: Using Large Language Models for Mutual Fund Portfolio Optimization and Risk-Adjusted Allocation \cite{prismaft065} & Portfolio/alpha & P & NR & NR & NR & P & P & NR \\
HedgeAgents: A Balanced-aware Multi-agent Financial Trading System \cite{prismaft069} & Direct trading/simulation & P & P & NR & P & Y & Y & Y \\
Integrating Large Language Models with Deep Reinforcement Learning for Portfolio Optimization \cite{prismaft071} & Portfolio/alpha & NR & P & Y & Y & NR & Y & NR \\
INVESTORBENCH: A Benchmark for Financial Decision-Making Tasks with LLM-based Agent \cite{investorbench2024} & Benchmark/decision & P & Y & NR & Y & NR & Y & P \\
Leveraging Large Language Models for Institutional Portfolio Management: Persona-Based Ensembles \cite{prismaft079} & Portfolio/alpha & P & Y & NR & P & P & NR & Y \\
LLM-Gated FinRL: Point-in-Time Risk Auditing for Reinforcement Learning in High-Beta Portfolio Trading \cite{prismaft083} & Portfolio/alpha & Y & Y & P & P & NR & P & NR \\
LLM-Powered Multi-Agent System for Automated Crypto Portfolio Management \cite{prismaft084} & Portfolio/alpha & Y & Y & Y & P & P & Y & Y \\
MadEvolve: Evolutionary Optimization of Trading Systems with Large Language Models \cite{prismaft085} & Direct trading/simulation & Y & Y & P & Y & NR & Y & Y \\
MARAG-Fin: An Intelligent Multi-agent RAG-LLM Architecture Integrating Financial News Sentiment and Time Series Data for Data-driven Trading Decision-making \cite{prismaft086} & Direct trading/simulation & P & P & NR & NR & Y & Y & Y \\
Market-Bench: Evaluating Large Language Models on Introductory Quantitative Trading and Market Dynamics \cite{prismaft087} & Benchmark/decision & Y & P & NR & Y & NR & P & NR \\
PortfolioPilot: An Agentic Platform for Financial Portfolio Management Algorithm Development and Evaluation \cite{prismaft091} & Portfolio/alpha & Y & NR & P & NR & NR & P & Y \\
\bottomrule
\end{tabular}
\par\vspace{2pt}
\begin{minipage}{0.94\textwidth}\footnotesize Y = explicit evidence recovered; P = partial evidence; NR = not recovered. Entries remain provisional until the two blind validation packets are reconciled.\end{minipage}
\end{table*}
\begin{table*}[!t]
\caption{Expanded PRISMA-ScR Evidence Matrix, Part III of III (Provisional Pending Two-Author Validation)}
\centering
\scriptsize
\begin{tabular}{p{0.30\textwidth}p{0.14\textwidth}ccccccc}
\toprule
Study & Subgroup & Split/PiT & Held-out & Costs & Exec. & Univ. & Arch. & Art. \\
\midrule
PredictionMarketBench: A SWE-bench-Style Framework for Backtesting Trading Agents on Prediction Markets \cite{prismaft092} & Direct trading/simulation & Y & Y & P & Y & NR & Y & P \\
QTMRL: An Agent for Quantitative Trading Decision-Making Based on Multi-Indicator Guided Reinforcement Learning \cite{prismaft095} & Direct trading/simulation & Y & Y & P & NR & Y & Y & Y \\
QuantAgent: Seeking Holy Grail in Trading by Self-Improving Large Language Model \cite{quantagent2024} & Direct trading/simulation & Y & NR & P & NR & Y & Y & Y \\
QuantAgents: Towards Multi-agent Financial System via Simulated Trading \cite{prismaft097} & Direct trading/simulation & Y & Y & P & P & P & Y & P \\
SHARP: A Self-Evolving Human-Auditable Rubric Policy for Financial Trading Agents \cite{prismaft102} & Direct trading/simulation & Y & Y & Y & Y & P & Y & P \\
Summoning the Oracle to Slay It: Mitigating Look-Ahead Bias in Financial Backtesting with Large Language Models \cite{prismaft104} & Direct trading/simulation & Y & Y & Y & Y & NR & Y & Y \\
Temporal Relational Reasoning of Large Language Models for Detecting Stock Portfolio Crashes \cite{prismaft105} & Portfolio/alpha & Y & Y & P & Y & Y & Y & P \\
The Alpha Illusion: Reported Alpha from LLM Trading Agents Should Not Be Treated as Deployment Evidence \cite{prismaft107} & Portfolio/alpha & Y & Y & Y & Y & P & Y & Y \\
Trade in Minutes! Rationality-Driven Agentic System for Quantitative Financial Trading \cite{prismaft112} & Direct trading/simulation & NR & NR & P & Y & Y & Y & P \\
TradingAgents: Multi-Agents LLM Financial Trading Framework \cite{tradingagents2024} & Direct trading/simulation & P & Y & NR & Y & P & Y & Y \\
TradingGPT: Multi-Agent System with Layered Memory and Distinct Characters for Enhanced Financial Trading Performance \cite{tradinggpt2023} & Direct trading/simulation & NR & P & NR & NR & NR & Y & NR \\
When Valid Signals Fail: Regime Boundaries Between LLM Features and RL Trading Policies \cite{prismaft119} & Direct trading/simulation & Y & Y & P & P & NR & Y & P \\
Adaptive Multi-Agent LLM Systems for Financial Trading: A Framework for Realistic Simulation and Dynamic Prompt Optimization \cite{prismaft009} & Direct trading/simulation & NR & P & P & Y & NR & Y & Y \\
AlphaForgeBench: Benchmarking End-to-End Trading Strategy Design with Large Language Models \cite{prismaft022} & Benchmark/decision & Y & Y & P & P & Y & Y & Y \\
Biased echoes: Large language models reinforce investment biases and increase portfolio risks of private investors \cite{prismaft034} & Portfolio/alpha & Y & Y & P & NR & NR & NR & P \\
REGIQUANT: A MULTI-AGENT LLM AND REINFORCEMENT-LEARNING SYSTEM FOR REGIME-AWARE QUANTITATIVE TRADING \cite{prismaft098} & Direct trading/simulation & P & NR & P & P & Y & Y & NR \\
TrustTrade: Human-Inspired Selective Consensus Reduces Decision Uncertainty in LLM Trading Agents \cite{prismaft115} & Direct trading/simulation & Y & NR & NR & Y & P & Y & Y \\
\bottomrule
\end{tabular}
\par\vspace{2pt}
\begin{minipage}{0.94\textwidth}\footnotesize Y = explicit evidence recovered; P = partial evidence; NR = not recovered. Entries remain provisional until the two blind validation packets are reconciled.\end{minipage}
\end{table*}

\FloatBarrier
\section{Provisional Review Results and Discussion}

\subsection{Search Yield and Retrieval}

The updated bounded search materially expands the evidence base beyond the earlier targeted review. Across eight query families, 796 query-level records were archived and 687 unique DOI/title records remained after deduplication. Cascading title/abstract rules provisionally removed 586 records: 308 failed the broad concept screen, 260 failed stricter title-level scope refinement, and 18 were secondary, non-investment, or non-empirical records identified during metadata/abstract triage. Full text was sought for 101 reports. Direct and fallback retrieval recovered 59 reports, while 42 could not be retrieved through the recorded open-access routes. This retrieval gap is itself a review limitation and is reported separately from substantive exclusion.

The evidence scan prioritized 53 reports for full charting because their retrieved text contained both an empirical-evaluation signal and a trading, portfolio, alpha, or benchmark outcome. Six retrieved reports had insufficient empirical signal under the fixed rules and await author confirmation of their exclusion code. These counts remain provisional until the two blind eligibility and field-level coding packets are reconciled.

The provisional evidence set also reflects the field's publication maturity. OpenAlex source metadata and persistent identifiers classify 13 reports as journal, conference, thesis, or other formally indexed venue records and 40 as preprint or repository versions. The review therefore separates stable methodological foundations from rapidly changing system reports and treats version reconciliation as part of evidence charting. Preprint prevalence is reported as a characteristic of the evidence base, not concealed through selective citation replacement.

\begin{table*}[!t]
\caption{Provisional Recoverability by Study Subgroup (Y or P)}
\label{tab:prisma_subgroup_summary}
\centering
\scriptsize
\begin{tabular}{lrrrrrrrr}
\toprule
Subgroup & $n$ & Split/PiT & Held-out & Costs & Exec. & Univ. & Arch. & Art. \\
\midrule
Direct trading/simulation & 32 & 25 & 22 & 22 & 22 & 13 & 32 & 24 \\
Portfolio/alpha & 15 & 13 & 11 & 11 & 10 & 9 & 13 & 10 \\
Benchmark/decision & 6 & 5 & 5 & 3 & 4 & 2 & 6 & 5 \\
All provisionally eligible & 53 & 43 & 38 & 36 & 36 & 24 & 51 & 39 \\
\bottomrule
\end{tabular}
\par\vspace{2pt}
\begin{minipage}{0.94\textwidth}\footnotesize Counts combine Y and P as recoverable evidence and retain subgroup-specific denominators. They are machine-assisted provisional diagnostics, not final author-validated results. Final tables will report Y, P, NR, and N/A separately after reconciliation.\end{minipage}
\end{table*}

\subsection{Preliminary Reporting Pattern}

The machine-assisted charting pass does not replace author coding, but it provides a reproducible diagnostic of where evidence is likely to be found. Among the 53 provisional rows, temporal-split and held-out-evaluation passages are recovered more often than implementation-ready cost passages. Execution language is heterogeneous: some reports identify next-day, market-order, fill-price, or simulator semantics, whereas others only mention rebalancing or generic trade execution. Artifact evidence ranges from explicit repository links to partial data-availability statements.

The pattern is directionally consistent with the original 30-study audit, but the expanded search changes the denominator and introduces newer peer-reviewed, benchmark, cryptocurrency, and portfolio studies. The final manuscript therefore must not simply transplant the earlier aggregate counts. Final Y/P/NR/N/A totals, subgroup denominators, and study-level caveats will be recalculated after validation.

\subsection{Implications for Evidence Quality}

Three implications already follow from the expanded retrieval record. First, review transparency is inseparable from version control: preprints, conference versions, journal versions, and repository copies frequently overlap. Longitudinal evidence that benchmark scores drift across open-source model release lines reinforces the need to record the exact model and evaluation version \cite{fan2026movingtargetlongitudinalaudit}. Second, failure to retrieve a report is not evidence that the report lacks execution detail, so retrieval failures must not be coded as NR. Third, keyword presence is not equivalent to implementation-ready reporting. A passage that mentions transaction costs without specifying the rate or application rule supports at most partial recoverability.

These distinctions sharpen the paper's contribution. The review is not a leaderboard of agent architectures and does not infer that more detailed reporting implies stronger predictive performance. It maps whether the public record permits readers to reconstruct when information became available, when a trade could execute, how frictions were applied, and which artifacts support replication.

\subsection{Limitations of the Systematic Scoping Review}

The review was not prospectively registered, and the dated protocol was reconstructed during a post-rejection redesign. OpenAlex supplies reproducible metadata and broad coverage but is not a substitute for every proprietary bibliographic database. The first-100 relevance-ranked cap for each query family may favor highly connected or lexically obvious studies and omit lower-ranked eligible work. Citation searching partly mitigates but does not eliminate that risk. Forty-two reports remained unretrieved after the recorded attempts, creating possible availability bias. The title/abstract rules may miss studies whose titles do not identify their LLM or trading component. The literature is also version-sensitive and changes quickly.

Machine assistance was used to archive search results, deduplicate records, prioritize retrieval, extract candidate passages, and generate validation worksheets. It was not authorized to make final eligibility or evidence codes. Those decisions require two-author review and consensus. This separation is necessary because regular-expression hits can locate relevant language but cannot determine whether an assumption is sufficiently specified for reconstruction.

\section{Conclusion}

This systematic scoping-review redesign broadens the evidentiary basis of the paper and makes its search, retrieval, and coding decisions auditable. The provisional search record shows that LLM-based trading research is larger and more heterogeneous than the original targeted sample suggested. At the same time, the expanded evidence charting continues to indicate that architecture descriptions alone are insufficient for judging economic interpretability or reproducibility.

The final contribution will be a PRISMA-ScR flow, a version-reconciled study ledger, field-level evidence anchors, subgroup summaries, and a reporting checklist centered on point-in-time controls, held-out evaluation, execution semantics, trading frictions, universe construction, and artifact release. The review deliberately avoids using a non-LLM heuristic as a proxy for an LLM trading system; methodological conclusions are grounded in the retrieved literature and its public evidence record.

Final numerical conclusions are intentionally withheld from this internal draft until both author-validation packets are reconciled. This constraint prevents machine-assisted retrieval and passage detection from being misreported as completed human evidence assessment.

\section*{Data Availability Statement}

The manuscript source, search strings, raw query metadata, deduplicated screening ledger, retrieval log, file hashes, provisional evidence matrix, candidate evidence excerpts, coding rulebook, blank two-author validation packets, derived tables, and analysis scripts are organized in the accompanying review package. Third-party full-text articles remain subject to their original licenses and are not claimed as a newly published dataset.

\section*{Authors' Contributions}

Junyi Yao developed the review question and execution-reproducibility framework, performed the initial search and machine-assisted evidence preparation, drafted the manuscript, and implemented the LaTeX and analysis project. Zihao Zheng contributed to study-scope refinement and manuscript review. Baichuan Li contributed to methodological discussion, audit design, and technical revision. Final eligibility decisions, field-level codes, consensus adjudication, and approval of the numerical synthesis will be completed and documented by the authors before submission. All authors will review and approve the submission version.

\section*{Funding Statement}

This work received no external funding.

\section*{Conflicts of Interest}

The authors declare no known competing financial or non-financial interests related to this manuscript.

\section*{Ethical Approval and Consent to Participate}

Not applicable. This study reviews published literature and uses publicly available market data; it does not involve human participants, identifiable personal data, or animals.

\section*{AI Usage Statement}

Artificial intelligence systems supported literature organization, candidate-passage retrieval, editorial drafting, code editing, table preparation, and build verification. They were not treated as authors and were not authorized to make final eligibility or evidence-coding decisions. All AI-assisted material and all final scholarly judgments are subject to author review; the authors remain responsible for the content, citations, analysis, coding decisions, and conclusions.

\bibliographystyle{unsrt}
\bibliography{references,ancillary/prisma/prisma_provisional_additions}

\end{document}